\documentclass{article}
\usepackage[nonatbib, preprint]{neurips_2023}
\usepackage[utf8]{inputenc} 
\usepackage[T1]{fontenc}    
\usepackage{hyperref}       
\usepackage{url}            
\usepackage{booktabs}       
\usepackage{amsfonts}       
\usepackage{nicefrac}       
\usepackage{microtype}      
\usepackage{xcolor}         
\usepackage{graphicx}
\usepackage{comment}
\usepackage{multicol}
\usepackage{color}
\usepackage{tabularray}
\usepackage{adjustbox}
\usepackage{multirow}
\usepackage{array} 
\usepackage{multirow} 
\usepackage{adjustbox} 
\usepackage[draft]{pgf}
\usepackage{comment}
\usepackage{float}
\usepackage{amsfonts, amsmath, amsthm, amssymb}
\usepackage{siunitx}
\title{New Benchmarks for Asian Facial Recognition Tasks: Face Classification with Large Foundation Models}

\author{Jinwoo Seo$^{*1}$ \; Soora Choi$^{*2}$ \; Eungyeom Ha$^{3}$ \; Beomjune Kim$^4$ \; \stepcounter{footnote}Dongbin Na$^{5}$\thanks{Correspondence to dongbinna@postech.ac.kr} \\ $^1$Catholic University \; $^2$Chung-Ang University \; $^3$Yonsei University \; $^4$SEOULTECH \; $^5$POSTECH}

\begin{document}

\def\thefootnote{*}\footnotetext{These authors contributed equally to this work.}

\maketitle

\begin{abstract}

The face classification system is an important tool for recognizing personal identity properly.
This paper introduces a new Large-Scale Korean Influencer Dataset named \textit{KoIn}.
Our presented dataset contains many real-world photos of Korean celebrities in various environments that might contain stage lighting, backup dancers, and background objects.
These various images can be useful for training classification models classifying K-influencers.
Most of the images in our proposed dataset have been collected from social network services (SNS) such as Instagram.
Our dataset, \textit{KoIn}, contains over 100,000 K-influencer photos from over 100 Korean celebrity classes.
Moreover, our dataset provides additional \textit{hard case} samples such as images including human faces with masks and hats.
We note that the \textit{hard case} samples are greatly useful in evaluating the robustness of the classification systems.
We have extensively conducted several experiments utilizing various classification models to validate the effectiveness of our proposed dataset.
Specifically, we demonstrate that recent state-of-the-art (SOTA) foundation architectures show decent classification performance when trained on our proposed dataset.
In this paper, we also analyze the robustness performance against \textit{hard case} samples of large-scale foundation models when we fine-tune the foundation models on the \textit{normal cases} of the proposed dataset, \textit{KoIn}.
Our presented dataset and codes will be publicly available at \textcolor{blue}{\url{https://github.com/dukong1/KoIn_Benchmark_Dataset}}.

\end{abstract}

\section{Introduction}

Facial recognition is a crucial tool that could be used for facial identification to authorize appropriate users by utilizing the feature representations of facial images.
Especially, facial classification systems have been largely adopted in various industries due to their convenience compared to conventional identification methods such as fingerprint recognition, retina recognition, and personal passwords~\cite{face_recognition, biometic_recognition, biometic_recog2}.

Recently, in the computer vision domain, deep convolutional neural networks (CNNs) have produced state-of-the-art (SOTA) detection performance for various tasks~\cite{facial_cnn, resnet, densenet, efficientnet, googlenet, mobilenet}.
This work focuses on the CNN-based facial classification system recognizing facial features.
The CNN-based face recognition models have been broadly used in various applications including airport immigration~\cite{airport_recog, air_rec, air_rec2}, online banking~\cite{banking_net, bank_rec}, cyber-security~\cite{cyber1, cyber2}, phishing account detection~\cite{phishing1, phishing_rec}, and medical services~\cite{medical1, medical_recog}.
The recent remarkable success of deep-learning applications is notably attributed to the large and high-fidelity dataset.
For example, the generally used image classification dataset ImageNet-1k~\cite{imagenet-1k} has about 1,000 images per class across 1,000 classes.
For the facial classification system, various facial image datasets are also presented in previous studies~\cite{celeba}.
However, the existing face datasets have limitations in that the number of samples per class is relatively small, sometimes insufficient for training and evaluating the facial classification models.
For example, the recent CelebA-HQ Facial Identity Recognition Dataset~\cite{celeba-HQ} contains about 20 images for each person, which is not enough to evaluate the classification performance of a model.
Furthermore, most existing facial image datasets mainly include Caucasians~\cite{white_races}.
In contrast, the number of image samples for Asian celebrities is relatively small, such as K-pop singers and famous Asian celebrities.

\begin{figure}[H]
    \centering
    \includegraphics[width=0.65\linewidth]{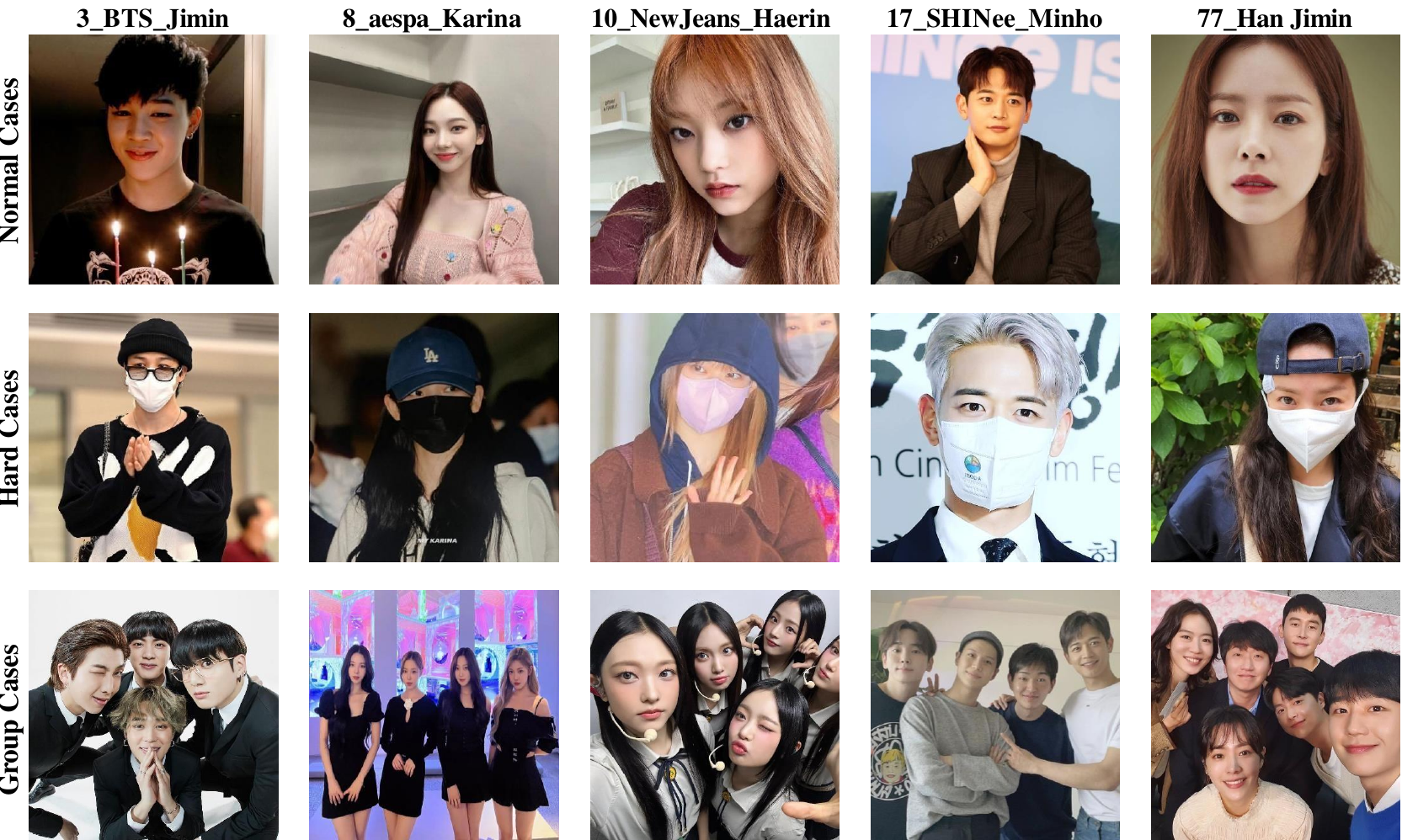}
    \caption{The showcases of image samples in our \textit{KoIn}. The first row shows \textit{normal case} images. The second row shows \textit{hard case} examples with masked faces. The last row shows different \textit{hard case} examples representing a group of people.}
    \label{fig:KI}
\end{figure}

To remedy this issue, we propose a new large-scale Korean influencer dataset, KoIn, that contains more than 1,000 real-world images per celebrity.
To the best of our knowledge, the large-scale facial recognition dataset whose size of each class is more than 1,000 for the classification system has not been presented yet.
Constructing high-fidelity image datasets is challenging because real-world images are affected by complex factors such as image post-processing, camera lighting, and the background, including backup dancers and stage steam.
With our extensive efforts, we present a large-scale Korean influencer dataset named \textbf{KoIn}, which contains 100 influencer classes (categories).
We have collected face images from a variety of websites, including social network services (SNS), Google, and other online services.
As a result, we have constructed the whole dataset that contains more than 100,000 images by collecting and curating more than 1,000 images for each category.
To the best of our knowledge, the \textit{KoIn} is the most large-scale Asian face classification dataset including many Korean celebrities, which can be greatly useful as a benchmark for developing facial recognition systems.
We illustrate some data examples of our proposed dataset in Figure~\ref{fig:KI}.
For evaluating the effectiveness of our dataset, we have utilized and evaluated individual state-of-the-art deep-learning models, including ResNet~\cite{resnet}, DenseNet~\cite{densenet}, and EfficientNet~\cite{efficientnet}. We have also leveraged the few-shot learning using recent SOTA foundation models such as BiT~\cite{BiT} and LIP~\cite{CLIP}.
Our experimental results have shown that our \textit{KoIn} is greatly useful in evaluating the robustness of the various facial classification models, including large foundation models.

The main contributions of our work are as follows:
\begin{itemize}
   
  \item We publicly present \textit{KoIn}, a large-scale Korean influencer facial image dataset with more than 100,000 images across 100 classes, which is suitable for training and evaluating the facial classification baseline models as a new benchmark facial recognition dataset.

 \item We provide variations of the proposed dataset, which are \textit{KoIn100}, \textit{KoIn50}, and \textit{KoIn10}. These datasets contain 100, 50, and 10 personal identities, respectively.

  \item We have trained the SOTA baseline CNN architectures and leveraged large foundation models. We have applied transfer-learning methods to train and evaluate the facial classification models. Whole trained models, source codes, and datasets are publicly available. We hope our presented resources will be useful for the field of facial recognition research.
\end{itemize}

This paper consists of the following sections.
Section~\ref{section2} briefly reviews the previously presented datasets and state-of-the-art face recognition methods.
Section~\ref{section3} addresses the collection and labeling procedure for constructing our \textit{KoIn} dataset.
In Section~\ref{section4}, we evaluate the usefulness of our proposed dataset by utilizing several benchmark deep-learning models and report the classification performance of the trained models.
In Section\ref{section5}, we summarize the proposed dataset and methods and result in the conclusion.

\section{Background and Related Work}

\label{section2}

\subsection{Deep Learning-based Image Classification and Transfer Learning}

The deep learning-based classification models have produced notable success in the various image recognition domains~\cite{resnet,densenet,efficientnet,alexnet}. For example, CNN-based architectures have achieved better performance than humans~\cite{vggnet} in the well-known image classification benchmark dataset~\cite{imagenet-1k}.
For improving the classification performance, the transfer-learning approach has been largely adopted~\cite{transfer_learning}. The general usage of transfer learning is \textit{fine-tuning}, which utilizes the foundation models and updates the weights of the foundation models to train on our dataset, whose size is generally small.
Especially, BiT~\cite{BiT} shows considerably superior performance in various image classification tasks. For example, fine-tuning the BiT foundation model on only 1,000 images can significantly surpass the relatively small baseline models trained on the 1,000,000 images from scratch in various benchmarks.
Therefore, many image recognition applications adopt and utilize transfer-learning methods due to their effectiveness and convenience.~\cite{millet_crop_transfer_learning, weak_machine_transfer_learning, welding_defects_transfer_learning,bearing_fault_transfer_learning}
However, we note that the analysis of the effect of fine-tuning for robustness has not been broadly studied yet.
Therefore, in this work, we report the fine-tuning performance of the foundation models, including BiT~\cite{BiT} and CLIP~\cite{CLIP}, on our proposed dataset with extensive experiments.

\subsection{Celebrity Facial Dataset for Facial Recognition}

The face classification models have been largely used in various industry domains, especially for user authorization.
For example, a security service requires a facial image for entry~\cite{arcface, LLWT15, celeba}.
The most famous facial dataset, CelebA~\cite{celeba}, has been largely adopted for evaluating face recognition systems.
However, this dataset only contains a few images per person, which can result in relatively poor classification performance because most person categories have only below 100 images.
Other previous studies also present LFW~\cite{LFW}, YouTube Face Database (YFD)~\cite{YFD}, CelebFaces+~\cite{CelebFaces+}, and CASIA-WebFace~\cite{CASIA-WebFace}.
However, these datasets have limitations in that the number of images per class is small.
For example, LFW~\cite{LFW} contains 13,000 images of celebrity faces from its website environment.
Moreover, although the CelebFace+~\cite{CelebFaces+} dataset contains a total of 202,599 facial images, however, there are 10,177 celebrity classes.
Similarly, CASIA-WebFace~\cite{CASIA-WebFace} addresses about 10,000 celebrities. 
However, this dataset contains only 50,000 images.
We also note that the previously proposed datasets largely depend on Caucasians rather than Asian people~\cite{white_races}.
Therefore, our work introduces a new large dataset containing many Korean celebrity images.
Our proposed dataset contains more than 1,000 images for each person.
In addition, the large number of samples in our work includes not only individual views of a person but also group shots of three or more people.
To validate the effectiveness of our proposed dataset, we have trained and evaluated various classification models.
Especially, we have leveraged the transfer-learning methods with foundation models and evaluated the trained models on our novel \textit{hard case} and \textit{group case} samples.
We demonstrate our proposed dataset can be greatly useful in evaluating the generalization performance and robustness of the facial recognition models and also report the detailed analysis.

\section{Proposed Datasets}

\label{section3}

For constructing our dataset that contains various Korean influencers, we mainly focus on the world-famous boy groups (e.g., BTS), girl groups (e.g., Aespa), solo singers (e.g., IU), and actors (e.g., Gong Yoo appearing in the drama "Dokkebi") who appeared in dramas or movies in South Korea.
In conclusion, we have selected 100 people among them through the curation procedure.
Our \textit{KoIn} dataset consists of more than 100,000 images of Korean influencers.
We have directly collected images from online websites by leveraging web crawling utilities.
Then, the annotators thoughtfully have labeled the images. 
Most collected samples are photographs taken in various environments and conditions.
Therefore, we have set labeling standards for effectively constructing datasets.
We largely divide them into three categories: \textit{normal cases}, \textit{hard cases}, and \textit{group cases}.
We set reasonable standards and largely divide them into these three categories: \textit{normal case}, \textit{hard case}, and \textit{group case}.
Our dataset configuration can be greatly useful in evaluating classification models' classification performance and robustness.
For example, the \textit{hard case} dataset contains various unrecognizable humans in the images, such as faces with masks and sunglasses.
We can evaluate the robustness of the face classification models on these \textit{hard cases} and \textit{group cases} of our \textit{KoIn} dataset, which provides a new research direction for improving the model generalization performance.

\subsection{Data Collection and Pre-processing}

In the data collection stage, we have collected most images from various SNS channels such as Instagram and online blogs.
Influencers tend to upload diverse photos to Instagram in a real-time manner worldwide.
Moreover, both the official accounts of the influencers and the accounts of their big fans generally contain a large amount of photos.
Thus, we also utilize these SNS accounts to minimize \textit{duplicated} images, non-celebrity photos, and photos without faces. Through this process, we finally have constructed an average of 1,000 pictures per class.
We note that the total number of categories of the proposed large-scale Korean influencer dataset (\textit{KoIn}) is 100.
Each image has only one label from 0 to 99 (multi-class classification), and at least 1,000 images are gathered for each class.
To reduce the unexpected biases of the proposed dataset, we divide the \textit{KoIn} dataset at a ratio of 50:50 for \textit{gender}, one of the most discriminative attributes for the face classification.
Moreover, we note that our proposed dataset only consists of highly curated images.
We then process all the images using JPG~\cite{opencv} compression libraries to constrain the size of a single image file not to exceed 1 MB.

\subsection{Data Categorization}

We have divided our whole collected dataset into three types of datasets using the following criterion.
We note that our novel \textit{hard case} images and \textit{group case} images are greatly useful in evaluating the robustness of the classification models.
We have annotated all the group case samples using the multi-label annotation setting.
If the image contains two or more people, we label all the target people in the annotation file.

\begin{itemize}
    \item \textbf{Hard cases}: We label an image as the \textit{hard case} sample if the person is wearing a mask or a hat. We also label the images if they cover their faces with their hands as \textit{hard case} images or if a side view is heavily covered.
    \item \textbf{Group cases}: We construct the \textit{group cases} samples by collecting K-pop boy and girl group images that contain at least four people. We set the standard to four people. We note that some images equal to or less than three people belong to the \textit{normal case} datasets.
    \item \textbf{Normal cases}: If the images do not belong to the categories above, we label the images as \textit{normal cases}. Most images are considered to be \textit{normal cases}. The number of \textit{normal cases} is relatively large compared to the \textit{hard case} and \textit{group case} samples.
\end{itemize}

\section{Experiments}

\label{section4}
We have conducted extensive experiments using our proposed \textit{KoIn} dataset. In this section, we first describe the experimental setup, introduce the experimental protocol, and provide the experimental results and analysis.

\subsection{Experimental Settings}

We have conducted all experiments using the PyTorch~\cite{pytorch} deep-learning framework.
In this paper, we analyze the classification performance of foundation models utilizing our proposed dataset.
Thus, we mainly use fine-tuning the foundation classification models on our dataset.
We adopt the stochastic gradient descent (SGD) optimization method with momentum.
In the training phase, the initial learning rate is set to 0.002 with a momentum of 0.9 for the transfer learning.
In the progression of training steps, We set the learning rate to decay to be 10\% of the previous learning rate every 5 epochs. 
The total number of epochs is 25 for the pre-trained models, and the total number of epochs is 200 for the models when training the models from scratch.

As the input pre-processing in the training phase, we first resize all images with random scaling to 256$\times$256 for baseline CNN-based architectures including ResNet, DenseNet, and EfficientNet. Then, we use 160$\times$160 for BiT-M, 336$\times$336 for ViT-L/14@336, and 224$\times$224 for all other ViT foundation models.
Then, we apply the \textit{Random Horizontal Flip} for data augmentation. 
We have used the pre-trained models trained on the ImageNet dataset~\cite{Imagenet-21k, JFT} and fine-tuned the networks with our \textit{KoIn} dataset except for the BiT models. 
In the validation and test phases, we just resize the image resolution and do not use other data augmentation methods. We have conducted all the experiments with a workstation that contains two GPU devices, T4 GPU and A100 GPU.

\subsection{Experimental Protocol}

We divide the whole dataset into training, validation, and test datasets by random selection. For the experiment section of this paper, we have experimented with the \textit{KoIn50}, containing over 53,556 training images. For training the models, we have adopted 10\% images among them as the validation dataset.
The test dataset consists of 5,000 images in which each class contains 100 images.
We have conducted comprehensive experiments using various popular deep-learning network architectures with different structures and different numbers of layers.
Specifically, we report the classification performance of ResNet (18, 34, 50, 101)~\cite{resnet}, DenseNet (121)~\cite{densenet}, EfficientNet (B0)~\cite{efficientnet}, BiT (M-R50, R101)~\cite{BiT}, and CLIP (RN50, RN101, ViT-B/32, ViT-L/14, ViT-L/14@336px)~\cite{CLIP} using our proposed dataset.
In addition, we also analyze the classification performance of ResNet18 and ResNet50 models~\cite{resnet} trained on our proposed dataset from scratch without pre-training.
We note that we also present a distinct \textit{hard case} evaluation dataset.
We evaluate the robustness of the models on \textit{hard case} samples, including masked faces and faces with glasses that are difficult to classify for the modern deep-learning models correctly.
Moreover, we also consider \textit{group photo images}, in which many people appear, as the \textit{hard case} samples.
If the model finds \textit{any} person in the group case, we consider that case \textit{correct}.
We expect that the \textit{group case} samples can also be used for evaluating the calibration performance of a classification model for future work~\cite{calibration}.
All experiments consistently adopt the same pre-processing and post-processing procedures for a fair comparison.
For example, we use the same input size and normalization layer for all the deep neural networks.

\subsection{Experimental results}

The classification performance of various deep networks is reported in Table~\ref{tab:KoIn_table1}.
We adopt both the top-1 and top-3 accuracy as evaluation metrics.
According to Table~\ref{tab:KoIn_table1}, the best top-ranked classification performance of the test dataset is 71.81\% obtained from DenseNet121, and the second-best result is derived from ResNet101, which achieves 69.41\%.
Table~\ref{tab:KoIn_table3} shows the classification performance of the transfer-learning foundation models.
We have experimented with transfer-learning methods using 5, 10, and 50 images per class, respectively.
BiT-M-R50 and R101 do not show significant differences compared to the performance of CNN models, but CLIP~\cite{CLIP} shows superior performance compared to other methods.
CLIP-RN50 achieves 39.82\% with only 5 images per class training, and ViT-L/14@336px shows an accuracy of 78.50\%, which is the best performance. 
The CLIP, devised based on zero-shot settings, can also perform greatly without additional fine-tuning.
We postulate the lower performance of BiT-M~\cite{BiT} is caused by the fact that BiT-M benchmark models are pre-trained on fewer human face images compared to CLIP, and the used benchmarks adopt relatively low-resolution input images.

\begin{table}[ht] 
    \renewcommand\arraystretch{1.0}
    \caption{Transfer-learning classification performance of various representative CNN-based image classification models fine-tuned on our \textit{KoIn} dataset: ResNet, DenseNet, EfficientNet. In the table, $\approx$1000 represents about 1,000 images per class, which indicates the case using our full dataset. The evaluation performance is reported using the \textit{normal case} test dataset.}
    \label{tab:KoIn_table1}
    \centering
    \begin{adjustbox}{width=9cm, center}
        \begin{tabular}{c|c|c|ccc}
            \hline
            \multirow{3}{*}{\textbf{Training Dataset}}\rule{0pt}{11pt}& \multirow{3}{*}{\textbf{\begin{tabular}[c]{@{}c@{}}Base Models Pre-trained \\ on ImageNet~\cite{imagenet-1k} \end{tabular}}}   
            & \multirow{3}{*}{\textbf{\begin{tabular}[c]{@{}c@{}}\# of Training \\ Images Per Class \end{tabular}}} 
            & \multicolumn{3}{c}{\textbf{Evaluation Scores (\%)}}\\ 
            \cline{4-6}\rule{0pt}{11pt}& & & \multicolumn{2}{c|}{\textbf{Top-1 Accuracy}} & \textbf{Top-3 Accuracy}\\ 
            \cline{4-6}\rule{0pt}{11pt}& & & \textbf{Validation} & \textbf{Test} & \textbf{Test}\\ 
            \hline
            \multirow{15}{*}{\textbf{Normal Cases}}\rule{0pt}{11pt}& \multirow{5}{*}{\textbf{ResNet101~\cite{resnet}}}
            & 10 & 25.99 & 12.51 & 27.93\\
            \rule{0pt}{11pt}& & 100 & 65.40 & 30.79 & 50.58\\
            \rule{0pt}{11pt}& & 500 & 76.80 & 52.34 & 71.08\\
            \rule{0pt}{11pt}& & $\approx$1000 & 80.93 & 69.41 & 83.97\\ 
            \cline{2-6}\rule{0pt}{11pt}& \multirow{5}{*}{\textbf{DenseNet121~\cite{densenet}}}    
            & 10 & 19.99 & 12.34 & 26.58\\
            \rule{0pt}{11pt}& & 100 & 61.00 & 31.78 & 52.22\\
            \rule{0pt}{11pt}& & 500 & 78.03 & 58.18 & 75.51\\
            \rule{0pt}{11pt}& & $\approx$1000 & 80.97 & 71.81 & 85.68\\ 
            \cline{2-6}\rule{0pt}{11pt}& \multirow{5}{*}{\textbf{EfficientNetB2~\cite{efficientnet}}}          
            & 10 & 23.99 & 8.80 & 21.28\\
            \rule{0pt}{11pt}& & 100 & 56.40 & 29.96 & 51.13\\
            \rule{0pt}{11pt}& & 500 & 75.67 & 52.52 & 72.11\\
            \rule{0pt}{11pt}& & $\approx$1000 & 81.51 & 68.30 & 83.95\\ 
            \hline
        \end{tabular}
    \end{adjustbox}
\end{table}

\begin{table}[ht]
    \renewcommand\arraystretch{1.0}
    \caption{Classification performance of the transfer-learning with foundation models: BiT-M models (R50, R101)~\cite{BiT} and CLIP models (RN50, RN101, ViT-B/32, ViT-L/14, ViT-L/14@336px)~\cite{CLIP} using the \textit{KoIn} dataset. The experiments have been conducted in the few-shot learning setting following the original purpose of BiT and CLIP models. In this table, the models are trained on the \textit{normal case} training dataset and evaluated on the \textit{normal case} test dataset. The CLIP-ViT-L/14@336px model shows superior classification performance using a high-resolution input layer, which might indicate the input resolution is a key factor for our proposed facial classification task.}
    \label{tab:KoIn_table3}
    \begin{adjustbox}{width=12cm,center}
        \begin{tabular}{c|cccccc}
            \hline
            \multirow{3}{*}{\textbf{Foundation Models}}\rule{0pt}{11pt}& \multicolumn{6}{c}{\textbf{Evaluation Scores for Normal Case Test Dataset (\%)}}\\ 
            \cline{2-7}\rule{0pt}{11pt}& \multicolumn{2}{c|}{\textbf{Trained on 5 Images Per Class}} & \multicolumn{2}{c|}{\textbf{Trained on 10 Images Per Class}} & \multicolumn{2}{c}{\textbf{Trained on 50 Images Per Class}}\\ 
            \cline{2-7}\rule{0pt}{11pt}& \multicolumn{2}{c}{\textbf{Top-1 Accuracy}} & \multicolumn{2}{c}{\textbf{Top-1 Accuracy}} & \multicolumn{2}{c}{\textbf{Top-1 Accuracy}} \\ 
            \hline
            \textbf{BiT-M-R50~\cite{BiT}}\rule{0pt}{11pt} & \multicolumn{2}{c}{8.78} & \multicolumn{2}{c}{11.28} & \multicolumn{2}{c}{24.97}\\ 
            \cline{1-1}
            \textbf{BiT-M-R101~\cite{BiT}}\rule{0pt}{11pt} & \multicolumn{2}{c}{9.14} & \multicolumn{2}{c}{13.59} & \multicolumn{2}{c}{27.55}\\ 
            \cline{1-1}
            \textbf{CLIP-RN50~\cite{CLIP}}\rule{0pt}{11pt} & \multicolumn{2}{c}{39.40} & \multicolumn{2}{c}{45.34} & \multicolumn{2}{c}{61.82}\\
            \cline{1-1}
            \textbf{CLIP-RN101~\cite{CLIP}}\rule{0pt}{11pt} & \multicolumn{2}{c}{45.84} & \multicolumn{2}{c}{53.62} & \multicolumn{2}{c}{68.70}\\
            \cline{1-1}
            \textbf{CLIP-ViT-B/32~\cite{CLIP}}\rule{0pt}{11pt} & \multicolumn{2}{c}{39.82} & \multicolumn{2}{c}{48.18} & \multicolumn{2}{c}{64.52}\\
            \cline{1-1}
            \textbf{CLIP-ViT-L/14~\cite{CLIP}}\rule{0pt}{11pt} & \multicolumn{2}{c}{71.34} & \multicolumn{2}{c}{79.24} & \multicolumn{2}{c}{88.40}\\
            \cline{1-1}
            \textbf{CLIP-ViT-L/14@336px~\cite{CLIP}}\rule{0pt}{11pt} & \multicolumn{2}{c}{78.50} & \multicolumn{2}{c}{84.86} & \multicolumn{2}{c}{91.70}\\
            \hline
        \end{tabular}
    \end{adjustbox}
\end{table}

Table~\ref{tab:KoIn_table4} and Table~\ref{tab:KoIn_table5} present the top-1 and top-3 accuracies on \textit{hard cases} and \textit{group cases} of the classification models that show relatively superior classification performance on \textit{normal cases}.
ResNet101 achieves an accuracy of 54.56\%, and DenseNet121 achieves an accuracy of 53.22\% for the \textit{hard cases}.
Interestingly, the large-scale foundation model CLIP-ViT-L/14@336px achieves 52.68\% for the \textit{hard cases} using just 5 images per class, which shows the superior robustness of the CLIP foundation model.
For the \textit{group cases}, this model achieves the best accuracy of 26.20\%, exceeding all other architectures.

\begin{table}[ht]
\renewcommand\arraystretch{1.0}
\centering
\caption{The robustness performance on \textit{hard case} and \textit{group case} samples of classical CNN-based classification models that show good classification performance on the \textit{normal cases}.}
\label{tab:KoIn_table4}
\begin{adjustbox}{width=6cm,center}
\begin{tabular}{c|c|cc}
\hline
\multirow{3}{*}{\textbf{Datasets}} & 
\multirow{3}{*}{\textbf{Architectures}}\rule{0pt}{11pt}& 
\multicolumn{2}{c}{\textbf{Evaluation Scores (\%)}}\\ 
\cline{3-4} 
& \rule{0pt}{11pt}& 
\multicolumn{1}{c|}{\textbf{Top-1 Accuracy}} & \textbf{Top-3 Accuracy}\\ 
\cline{3-4} 
& \rule{0pt}{11pt}& 
\multicolumn{1}{c}{\textbf{Test Accuracy}} & 
\textbf{Test Accuracy}\\ 
\hline
\multirow{3}{*}{\textbf{Hard Cases}}\rule{0pt}{11pt}& 
\multirow{1}{*}{\textbf{ResNet101}} 
& 54.56   & 73.06\\ 
\rule{0pt}{11pt}& \multirow{1}{*}{\textbf{DenseNet121}}
& 53.22 & 72.25\\ 
\rule{0pt}{11pt}& 
\multirow{1}{*}{\textbf{EfficientNetB2}} 
& 52.82 & 69.71\\
\hline
\multirow{3}{*}{\textbf{Group Cases}}\rule{0pt}{11pt}& 
\multirow{1}{*}{\textbf{ResNet101}}      
& 19.31 & 38.62\\ 
\rule{0pt}{11pt}& 
\multirow{1}{*}{\textbf{DenseNet121}}   
& 20.34 & 39.66\\ 
\rule{0pt}{11pt}& 
\multirow{1}{*}{\textbf{EfficientNetB2}} 
& 21.03 & 39.66\\
\hline
\end{tabular}
\end{adjustbox}
\end{table}

\begin{table}[ht]
\renewcommand\arraystretch{1.0}
\centering
\caption{The robustness performance on \textit{hard case} and \textit{group case} samples of the large foundation models that show superior classification performance on the \textit{normal cases}.}
\label{tab:KoIn_table5}
\begin{adjustbox}{width=12.0cm,center}
\begin{tabular}{c|c|ccc}
\hline
\multirow{3}{*}{\textbf{Datasets}}\rule{0pt}{11pt}& \multirow{3}{*}{\textbf{Architectures}}
& \multicolumn{3}{c}{\textbf{Evaluation Scores (\%)}}                                                      \\ \cline{3-5}\rule{0pt}{11pt}&
& \multicolumn{1}{c|}{\textbf{Trained on 5 Images Per Class}} & \multicolumn{1}{c|}{\textbf{Trained on 10 Images Per Class}} & \textbf{Trained on 50 Images Per Class} \\ \cline{3-5}\rule{0pt}{11pt}&
& \multicolumn{1}{c}{\textbf{Test Accuracy}}
& \multicolumn{1}{c}{\textbf{Test Accuracy}}
& \textbf{Test Accuracy}
\\ \hline
\multirow{3}{*}{\textbf{Hard Cases}}\rule{0pt}{11pt}& \textbf{CLIP-ViT-B/32} & 18.63 & 26.27 & 35.79 \\
\rule{0pt}{11pt}& \textbf{CLIP-ViT-L/14} & 42.49 & 52.94 & 59.51 \\
\rule{0pt}{11pt}& \textbf{CLIP-ViT-L/14@336px} & 52.68 & 60.72  & 68.23
\\ \hline
\multirow{3}{*}{\textbf{Group Cases}}\rule{0pt}{11pt}& \textbf{CLIP-ViT-B/32} & 11.72 & 15.86 & 23.79 \\
\rule{0pt}{11pt}& \textbf{CLIP-ViT-L/14} & 22.75 & 29.65 & 37.93 \\
\rule{0pt}{11pt}& \textbf{CLIP-ViT-L/14@336px} & 26.20 & 37.24 & 43.79 
\\ \hline
\end{tabular}
\end{adjustbox}
\end{table}

\section{Conclusions}

\label{section5}
This paper introduces a novel and large-scale dataset, \textit{KoIn}, that contains various face images of Korean influencers.
Our research goal is to distribute useful facial datasets focusing on Korean celebrities among Asian facial photographs for research purposes in the facial recognition research domain.
The \textit{KoIn} dataset can be potentially used for face recognition services, fraud detection systems, and other related services.
The proposed dataset has 100,000 images across 100 categories, mainly containing highly-curated photos among web images.
In addition, we train well-known CNN architectures, including ResNet50, Resnet101, and Densnet121 models on our 50 classes version KoIn dataset (\textit{KoIn50}), and analyze the experimental results in the transfer-learning setting.
We also demonstrate the few-shot learning methods using BiT and CLIP, which are representative large-scale foundation models.
We hope our \textit{KoIn} will be useful for related research fields.

\bibliographystyle{plain}
\bibliography{neurips_2023}

\newpage

\appendix

\section{Data Distribution} 
KoIn Dataset is a large-scale facial dataset that contains more than 100,000 Korean celebrity images. Images in this dataset include diverse facial expressions and human poses. Moreover, this dataset covers various backgrounds, strong stage lighting, stage smoke, and other noises. We also provide individual image datasets \textit{group case} test dataset and \textit{hard case} test dataset. The \textit{hard case} dataset includes the faces wearing hats or masks and glasses.

\begin{figure}[ht]
    \centering
    \includegraphics[width=0.8\linewidth]{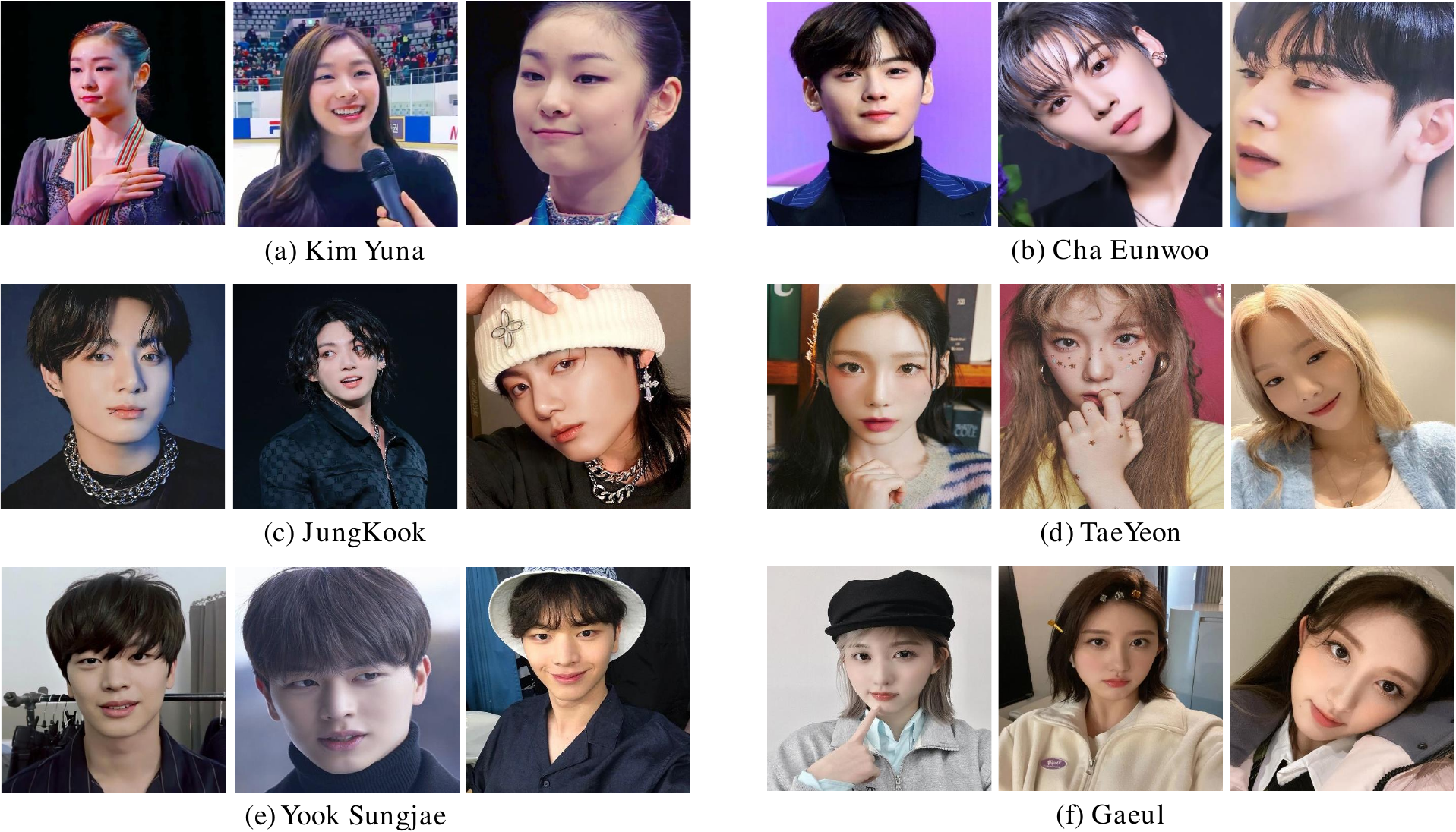}
    \caption{Example images of 6 Korean celebrities in the \textit{KoIn} dataset that has been used in the experiment section.}
    \label{fig:KI_detail}
\end{figure}

We have selectively chosen 50 celebrities and constructed the 53,556 images across 50 celebrities for constructing the \textit{KoIn50}.
Some of the images of the experiment are shown in Figure~\ref{fig:KI_detail}.
The index numbers and the number of images for the 50 classes are shown in Table~\ref{tab: Normal cases: Train dataset distribution}.
Furthermore, the detailed information of \textit{normal} and \textit{hard cases} are shown in Table~\ref{tab: Total images of the train, test dataset distribution}. 
After the paper is accepted, more detailed information and configurations on the entire dataset will be released as open-source.

Our whole KoIn100 dataset satisfies the following conditions:

\begin{itemize}
    \item 100 number of identities.
    \item 1,000 number of images per class.
    \item 100,000 number of the total of face images.
    \item 1,000 number of \textit{hard case} test images and 500 number of \textit{group case} test images.
\end{itemize}

\begin{table}[ht]
    \renewcommand\arraystretch{1.0}
    \caption{The distribution of training dataset used in the experiment section. We note that the training dataset only contains \textit{normal case} images.}
    \label{tab: Normal cases: Train dataset distribution}
    \centering
    \begin{adjustbox}{width=11.0cm,center}
        \begin{tabular}{c|cccccccccc}
            \hline
            \multirow{1}{*}{\textbf{Training Dataset}} 
            \rule{0pt}{11pt}& \multicolumn{10}{c}{\textbf{Number of Images Per Class}} \\ 
            
            \hline
            \multirow{13}{*}{\textbf{Normal Cases}}\rule{0pt}{11pt}& \textbf{0}  & \textbf{1}  & \textbf{2}  & \textbf{3}  & \textbf{4}  & \textbf{5}  & \textbf{6}  & \textbf{7}  & \textbf{8}  & \textbf{9}  \\ 
            \cline{2-11}\rule{0pt}{11pt}
            & 1,061 & 1,015 & 1,161 & 1,105 & 1,155 & 1,017 & 1,029 & 1,000 & 1001 & 1,025   \\ 
            \cline{2-11}\rule{0pt}{11pt}& \textbf{10} & \textbf{11} & \textbf{12} & \textbf{13} & \textbf{14} & \textbf{15} & \textbf{16} & \textbf{17} & \textbf{18} & \textbf{19} \\ 
            \cline{2-11}\rule{0pt}{11pt}
            & 1,264 & 1,055 & 1,028 & 1,071 & 1,034 & 1,162 & 1,118 & 1,026 & 1,145 & 1,133      \\ 
            \cline{2-11}\rule{0pt}{11pt}& \textbf{20} & \textbf{21} & \textbf{22} & \textbf{23} & \textbf{24} & \textbf{25} & \textbf{26} & \textbf{27} & \textbf{28} & \textbf{29} \\ 
            \cline{2-11}\rule{0pt}{11pt}
            & 1,077 & 1,154 & 1,243 & 1,077 & 1,037 & 1,068 & 1,038 & 1,011 & 1,002 & 1,115        \\ 
            \cline{2-11}\rule{0pt}{11pt}& \textbf{30} & \textbf{31} & \textbf{32} & \textbf{33} & \textbf{34} & \textbf{35} & \textbf{36} & \textbf{37} & \textbf{38} & \textbf{39} \\ 
            \cline{2-11}\rule{0pt}{11pt}
            & 1,001 & 1,001 & 1,058 & 1,092 & 1,105 & 1,093 & 1,033 & 1,016 & 1,153 & 1,041       \\ 
            \cline{2-11}\rule{0pt}{11pt}& \textbf{40} & \textbf{41} & \textbf{42} & \textbf{43} & \textbf{44} & \textbf{45} & \textbf{46} & \textbf{47} & \textbf{48} & \textbf{49} 
            \\ 
            \cline{2-11}\rule{0pt}{11pt}
            & 1,100 & 1,091 & 1,049 & 1,012 & 1,033 & 1,118 & 1,052 & 1,021 & 1,005 & 1,055         
            \\
            \hline
        \end{tabular}
    \end{adjustbox}
\end{table}

\begin{table}[ht]
    \renewcommand\arraystretch{1.0}
    \caption{The number of images in the training and test dataset of KoIn50 used in the experiment section.}
    \label{tab: Total images of the train, test dataset distribution}
    \centering
    \begin{adjustbox}{width=11cm,center}
        \begin{tabular}{c|c|c|c|c}
            \hline
            \rule{0pt}{11pt}& \multicolumn{1}{c|}{\textbf{Training Dataset}}
            & \multicolumn{3}{c}{\textbf{Test Dataset}} \\ 
            \cline{2-5}\rule{0pt}{11pt}& \textbf{Normal Cases (0-49)} 
            & \textbf{Normal Cases (0-49)} 
            & \textbf{Hard Cases} 
            & \textbf{Group Cases} \\
            \hline
            \textbf{Total Images}\rule{0pt}{11pt}& 53,556 & 5,000 & 752 & 290  \\ 
            \hline
        \end{tabular}
    \end{adjustbox}
\end{table}

\section{Case Annotation Guideline}
\begin{figure}[ht]
    \centering
    \includegraphics[width=0.75\linewidth]{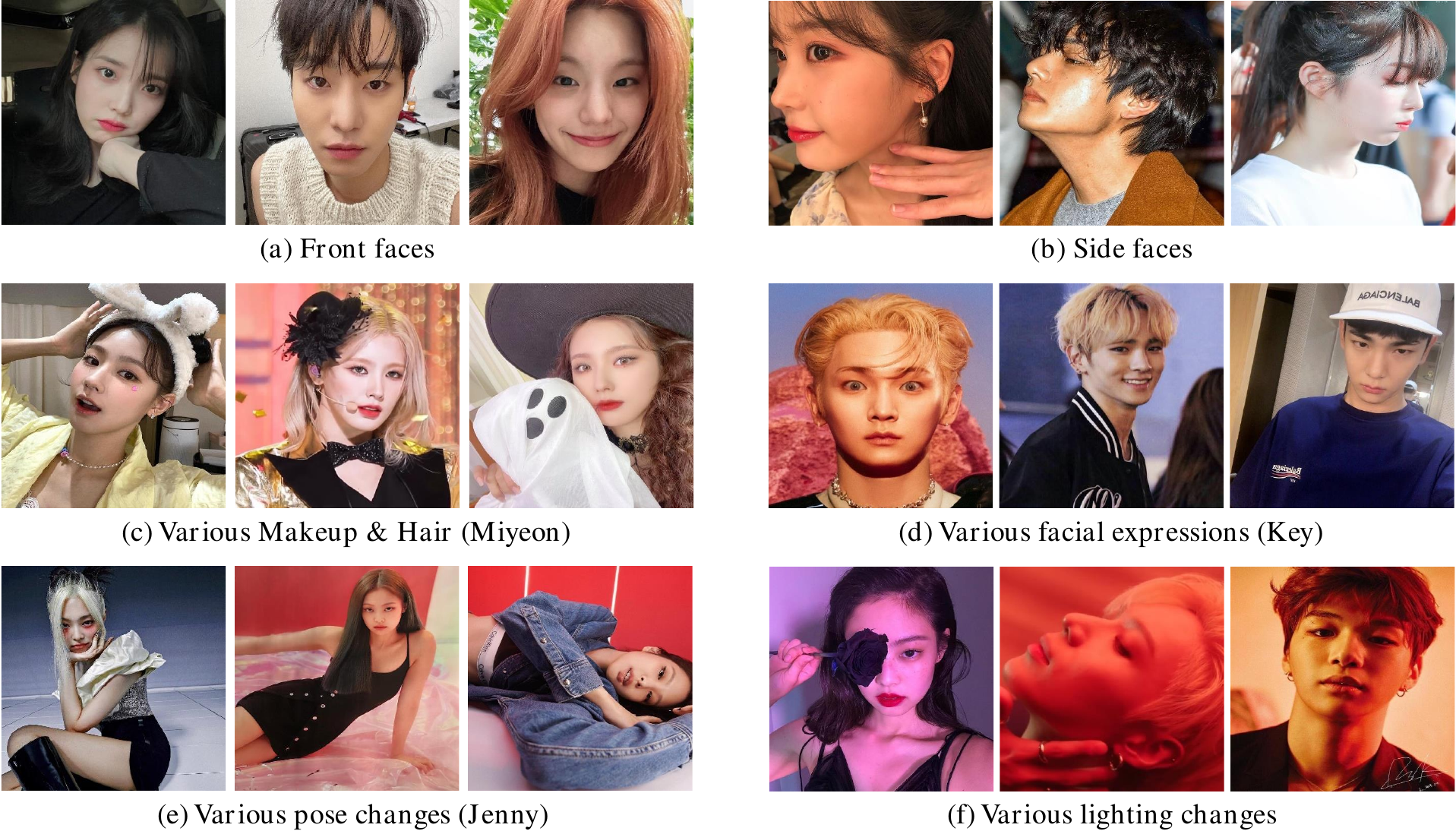}
    \caption{\textit{KoIn} dataset consists of various photos with (a) the front face, (b) the side face, and (c) a certain proportion of the face. As shown in the (d)-(f), our dataset includes various facial expression changes, hair \& make-up changes, and several environmental changes (e.g. lighting).}
    \label{fig:KI_detail2}
\end{figure}

For constructing the \textit{KoIn} dataset, the poor-quality or overlapping images are removed and we proceed annotation process as follows:
\begin{itemize}
    \item \textbf{Normal case}: \textit{KoIn} dataset includes photos with the front face, the side face, and a certain proportion of the face. The normal case dataset also contains the variation of facial expression, hair \& make-up, and stage lighting. 
    The examples are shown in Figure~\ref{fig:KI_detail2}
    \item \textbf{Hard case}: The \textit{hard case} dataset generally contains a face with a mask covering a large part of the face or with a hat.
    This dataset also includes various cases in which many parts of the face are cut off or tilted at an angle, which is difficult to distinguish.
    \item \textbf{Group case}: The \textit{group case} dataset contains many pictures that contain several celebrity people in a single photo
    Many singers belong to the team in South Korea.
    In an image of the group case dataset, at least one person is included among team members.
\end{itemize}

\section{Definition of Korean Influencer}

\begin{figure}[ht]
    \centering
    \includegraphics[width=0.8\linewidth]{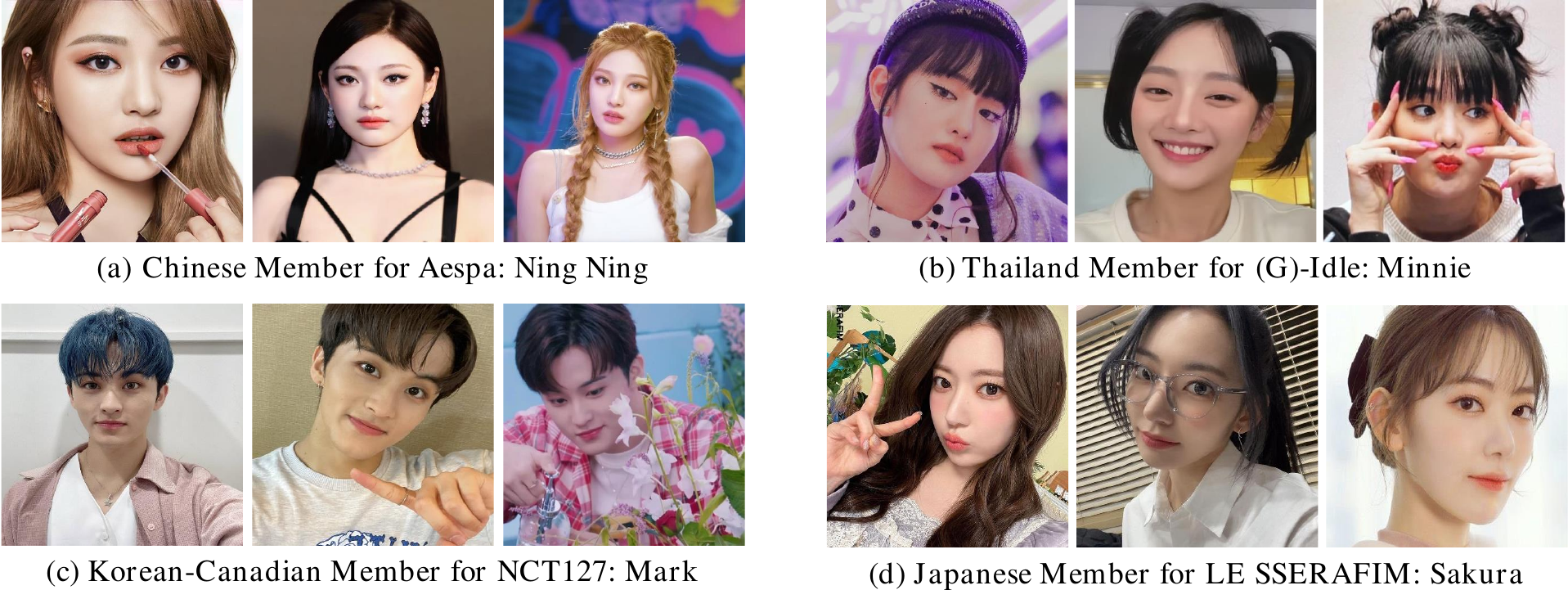}
    \caption{Example images of 4 foreign celebrities, debuted in Korea, from the \textit{KoIn} dataset used in the experiment.}
    \label{fig:KI_detail5}
\end{figure}

\textbf{Korean Influencer} denotes a person who debuted in South Korea in the entertainment industry. Debut usually goes through the trainee period of an entertainment agency, which is also the case when Korean sports stars enter the entertainment industry after retirement. A large part of this dataset consists of Koreans.
However, our dataset also includes ethnic Koreans, such as Korean Americans, and those who are Chinese but made their debut in Korea after the official training. In these cases, they are all Asian. Specifically, the 12 categories out of 100 classes in the \textit{KoIn} dataset belong to these cases, accounting for about 12\% images. The examples are shown in Figure~\ref{fig:KI_detail5}.

\begin{table}[ht]
    \renewcommand\arraystretch{1.0}
    \caption{\textcolor{black}{Performance comparison of various representative CNN-based image classification models fine-tuned on our \textit{KoIn} dataset: ResNet, DenseNet, and EfficientNet. In the table, $\approx$1000 denotes about 1,000 images per class, using our full dataset. The evaluation performance is reported using the \textit{normal case} test dataset.}}
    \label{tab:KoIn_full_table}
    \centering
    \begin{adjustbox}{width=11.0cm,center}
        \begin{tabular}{c|c|c|ccc}
            \hline
            \multirow{3}{*}{\textbf{Training Dataset}}\rule{0pt}{11pt}& \multirow{3}{*}{\textbf{\begin{tabular}[c]{@{}c@{}}Base Models Pre-trained \\ on ImageNet~\cite{imagenet-1k} \end{tabular}}}   
            & \multirow{3}{*}{\textbf{\begin{tabular}[c]{@{}c@{}}\# of Images \\ Per Class \end{tabular}}} 
            & \multicolumn{3}{c}{\textbf{Evaluation Scores (\%)}}\\ 
            \cline{4-6}\rule{0pt}{11pt}& & & \multicolumn{2}{c|}{\textbf{Top-1 Accuracy}} & \textbf{Top-3 Accuracy}\\ 
            \cline{4-6}\rule{0pt}{11pt}& & & \textbf{Validation} & \textbf{Test} & \textbf{Test}\\ 
            \hline
            \multirow{30}{*}{\textbf{Normal Cases}}\rule{0pt}{11pt}& \multirow{4}{*}{\textbf{ResNet18~\cite{resnet}}} 
            & 10 & 17.99 & 10.57 & 23.00\\
            \rule{0pt}{11pt}& & 100 & 57.40 & 26.86 & 46.92\\
            \rule{0pt}{11pt}& & 500 & 70.03 & 46.86 & 66.15\\
            \rule{0pt}{11pt}& & $\approx$1000 & 75.29 & 61.60 & 78.98\\ 
            \cline{2-6}\rule{0pt}{11pt}& \multirow{4}{*}{\textbf{ResNet34~\cite{resnet}}}
            & 10 & 25.99 & 12.48 & 25.73 \\
            \rule{0pt}{11pt}& & 100 & 53.20 & 27.73 & 47.28\\
            \rule{0pt}{11pt}& & 500 & 71.91 & 49.72 & 68.51\\
            \rule{0pt}{11pt}& & $\approx$1000 & 78.07 & 66.07 & 81.53\\ 
            \cline{2-6}\rule{0pt}{11pt}& \multirow{4}{*}{\textbf{ResNet50~\cite{resnet}}}       
            & 10 & 19.99 & 11.48 & 25.35\\
            \rule{0pt}{11pt}& & 100 & 56.20 & 30.83 & 50.58\\
            \rule{0pt}{11pt}& & 500 & 75.47 & 54.27 & 72.19\\
            \rule{0pt}{11pt}& & $\approx$1000 & 80.41 & 69.35 & 84.05\\ 
            \cline{2-6}\rule{0pt}{11pt}& \multirow{4}{*}{\textbf{ResNet101~\cite{resnet}}}
            & 10 & 25.99 & 12.51 & 27.93\\
            \rule{0pt}{11pt}& & 100 & 65.40 & 30.79 & 50.58\\
            \rule{0pt}{11pt}& & 500 & 76.80 & 52.34 & 71.08\\
            \rule{0pt}{11pt}& & $\approx$1000 & 80.93 & 69.41 & 83.97\\ 
            \cline{2-6}\rule{0pt}{11pt}& \multirow{4}{*}{\textbf{DenseNet121~\cite{densenet}}}    
            & 10 & 19.99 & 12.34 & 26.58\\
            \rule{0pt}{11pt}& & 100 & 61.00 & 31.78 & 52.22\\
            \rule{0pt}{11pt}& & 500 & 78.03 & 58.18 & 75.51\\
            \rule{0pt}{11pt}& & $\approx$1000 & 80.97 & 71.81 & 85.68\\ 
            \cline{2-6}\rule{0pt}{11pt}& \multirow{4}{*}{\textbf{EfficientNetB0~\cite{efficientnet}}} 
            & 10 & 21.99 & 7.75 & 20.40\\
            \rule{0pt}{11pt}& & 100 & 57.60 & 29.56 & 50.08\\
            \rule{0pt}{11pt}& & 500 & 74.00 & 49.56 & 71.71\\
            \rule{0pt}{11pt}& & $\approx$1000 & 78.61 & 65.34 & 82.10\\ 
            \cline{2-6}\rule{0pt}{11pt}& \multirow{4}{*}{\textbf{EfficientNetB2~\cite{efficientnet}}}          
            & 10 & 23.99 & 8.80 & 21.28\\
            \rule{0pt}{11pt}& & 100 & 56.40 & 29.96 & 51.13\\
            \rule{0pt}{11pt}& & 500 & 75.67 & 52.52 & 72.11\\
            \rule{0pt}{11pt}& & $\approx$1000 & 81.51 & 68.30 & 83.95\\ 
            \hline
        \end{tabular}
    \end{adjustbox}
\end{table}

\section{Detailed CNN Model Performance Analysis}

The classification performance of various deep networks is shown in Table~\ref{tab:KoIn_full_table}.
We report both the top-1 and top-3 accuracy.
According to Table~\ref{tab:KoIn_full_table}, the best top-ranked performance in the validation set is 81.51\%, achieved by EfficientNetB2. 
The validation performance of ResNet101 and DenseNet121 is also very close to the best results. 
The best classification performance for the test dataset is 71.81\% obtained from DenseNet121, and the second-best result is obtained from ResNet101, which achieves 69.41\%. 
Deep CNN models generally achieve better performance~\cite{googlenet, Inception_v3}.
We note that the ResNet models consistently improve their classification performance when the number of layers increases using the same network architecture in our experiments.
For example, ResNet18 achieves an accuracy of 61.60\%, and the relatively deep ResNet architecture using 101 layers achieves approximately 8\% improvement compared to ResNet18 on both validation and test sets.
Table~\ref{tab:KoIn_table2} represents the training results with the randomly initialized ResNet18 and ResNet50 models that are not pre-trained on the ImageNet-1k~\cite{imagenet-1k}.
To compare with the classification performance of ImageNet-1k pre-trained models, we have trained the ResNet (18, 50) models, which are the most commonly used architectures in the computer vision research field, from scratch on our training dataset.
When comparing with the ResNet models not adopting pre-training, the performance gaps for ResNet18 and ResNet50 are 17\% and 14\%, respectively, indicating that the ImageNet-1k pre-trained model shows better performance.

\begin{table}[ht]
    \renewcommand\arraystretch{1.0}
    \caption{The experimental results using ResNet18 and ResNet50 models. In this table, the models have been trained on the whole \textit{normal case} training dataset from scratch and evaluated on the test dataset.}
    \label{tab:KoIn_table2}
    \begin{adjustbox}{width=14.0cm,center}
        \begin{tabular}{c|cccccc}
            \hline
            \multirow{3}{*}{\textbf{Architectures}}\rule{0pt}{11pt}& \multicolumn{6}{c}{\textbf{Evaluation Scores (\%)}}\\ 
            \cline{2-7}\rule{0pt}{11pt}& \multicolumn{2}{c|}{\textbf{Normal Case Test Dataset}} & \multicolumn{2}{c|}{\textbf{Hard Case Test Dataset}} & \multicolumn{2}{c}{\textbf{Group Case Test Dataset}}\\ 
            \cline{2-7}\rule{0pt}{11pt}& \textbf{Top-1 Accuarcy} & \textbf{Top-3 Accuarcy} & \textbf{Top-1 Accuarcy} & \textbf{Top-3 Accuarcy} & \textbf{Top-1 Accuarcy} & \textbf{Top-3 Accuarcy}\\ 
            \hline
            \textbf{ResNet-18 (No Pre-trained)}\rule{0pt}{11pt}& 44.24 & 65.06 & 24.40 & 46.38 & 13.10 & 26.21\\ 
            \cline{1-1}
            \textbf{ResNet-50 (No Pre-trained)}\rule{0pt}{11pt}& 34.64 & 55.13 & 15.55 & 33.91 & 11.38 & 24.83\\ 
            \hline
        \end{tabular}
    \end{adjustbox}
\end{table}

\section{Discussion}

\begin{figure}[ht]
    \centering
    \includegraphics[width=0.8\linewidth]{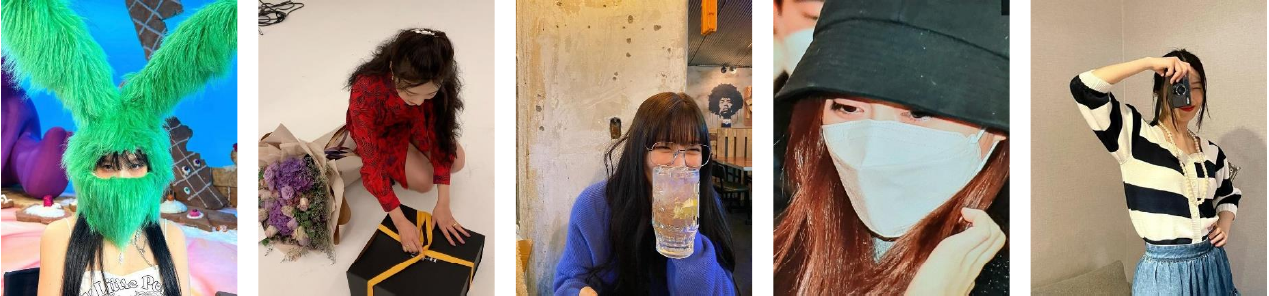}
    \caption{Examples of \textit{hard case} samples that were not properly classified even with the CLIP-ViT-L/14@336px model, which achieved the highest performance in our whole experiments.}
    \label{fig:KI_hard}
\end{figure}

In this work, we aim to construct an Asian face classification dataset that can be utilized to increase the robustness and classification performance in the face recognition tasks.
For the face classification tasks, the classification model needs to be able to classify face recognition as compliant performance, even in \textit{hard cases} and \textit{group cases}.
The ideal face recognition model can distinguish the presence or absence of a specific person even though various regions of the image are not easily discriminative.
Figure~\ref{fig:KI_hard} shows various \textit{hard case} images that were not distinguished by CLIP-ViT-L/14@336px model, which has demonstrated the best performance in the experiment.
The examples in Figure~\ref{fig:KI_hard} are representative images that should be classified only with information on the shape of eyes or information on the face angle different from the front angle of the face.
However, the image classification models trained on just \textit{normal cases} might not classify the \textit{hard cases} properly.
Therefore, for future work, we will continue collecting, training, and researching to develop accurate face classification models, that perform well even in more difficult cases, as shown in Figure~\ref{fig:KI_hard}.

\end{document}